\documentclass[10pt,twocolumn,letterpaper]{article}

\usepackage[pagenumbers]{cvpr} % To force page numbers, e.g. for an arXiv version

\usepackage{lineno} % required package, or else will raise error

\definecolor{cvprblue}{rgb}{0.21,0.49,0.74}
\usepackage[pagebackref,breaklinks,colorlinks,allcolors=cvprblue]{hyperref}

\usepackage{float}

\def\paperID{*****} % *** Enter the Paper ID here
\def\confName{CVPR}
\def\confYear{2025}

\title{
    \raisebox{-0.15\height}{\includegraphics[width=0.06\textwidth]{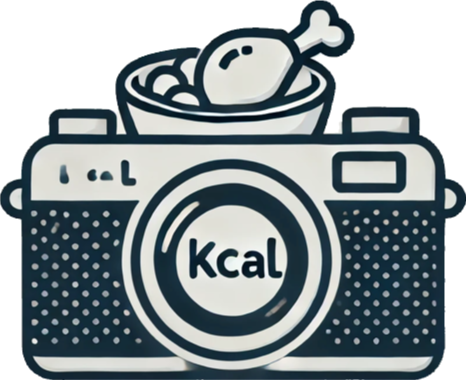}}  % Adjust width and raisebox as needed
    \textbf{\textit{CaLoRA}ify: 
    \textit{Ca}lorie Estimation with Visual-Text Pairing 
    \\ and \textit{LoRA}-Driven Visual Language Models}
}

\author{
    Dongyu Yao\textsuperscript{\dag} \quad Keling Yao\textsuperscript{\dag} \quad Junhong Zhou\textsuperscript{\dag} \quad Yinghao Zhang\textsuperscript{\dag} \\
    \textsuperscript{\dag}: Equal contribution\\
    Carnegie Mellon University
}

\begin{document}
\maketitle
\begin{abstract}
% The obesity phenomenon, known as the ``heavy" issue
The obesity phenomenon, known as the ``heavy" issue, is a leading cause of preventable chronic diseases worldwide. Traditional calorie estimation tools often rely on specific data formats or complex pipelines, limiting their practicality in real-world scenarios. Recently, vision-language models (VLMs) have excelled in understanding real-world contexts and enabling conversational interactions, making them ideal for downstream tasks such as ingredient analysis.  However, applying VLMs to calorie estimation requires domain-specific data and alignment strategies. To this end, we curated \textbf{CalData}, a 330K image-text pair dataset tailored for ingredient recognition and calorie estimation, combining a large-scale recipe dataset with detailed nutritional instructions for robust vision-language training. Built upon this dataset, we present \textbf{CaLoRAify}, a novel VLM framework aligning ingredient recognition and calorie estimation via training with visual-text pairs. In inference, users only need \textbf{ a single monocular food image} to estimate calories while retaining the flexibility of agent-based conversational interaction.
With Low-rank Adaptation (LoRA) and Retrieve-augmented Generation (RAG) techniques, our system enhances performance of foundational VLMs in the vertical domain of calorie estimation. Our code and data are fully open-sourced at \href{https://github.com/KennyYao2001/16824-CaLORAify}{https://github.com/KennyYao2001/16824-CaLORAify}.

\end{abstract}    
\section{Introduction}
\label{sec:intro}

% 1. Background and Motivation
Obesity has emerged as a significant public health crisis, affecting over 42.4\% of adults in the United States alone \cite{NIH}. It is ranked as the leading cause of preventable chronic diseases among developed nations.

Over the years, a variety of tools and methods have been developed to aid in calorie management, ranging from mobile applications to AI-powered systems. These tools have flooded the market, driven by the immense commercial value of addressing such a widespread issue. For instance, CalAI, a popular calorie estimation tool leveraging artificial intelligence, has recently reported annual revenues exceeding \$50 million, reflecting the increasing demand for accessible and accurate dietary management solutions \cite{CalAI_Revenue}. Similarly, platforms like MyFitnessPal \cite{MyFitnessPal_UserBase} and LoseIt \cite{LoseIt_Success} have garnered millions of users worldwide, further highlighting the growing market for such technologies.

\begin{figure}[H]
    \centering
    \includegraphics[width=\linewidth]{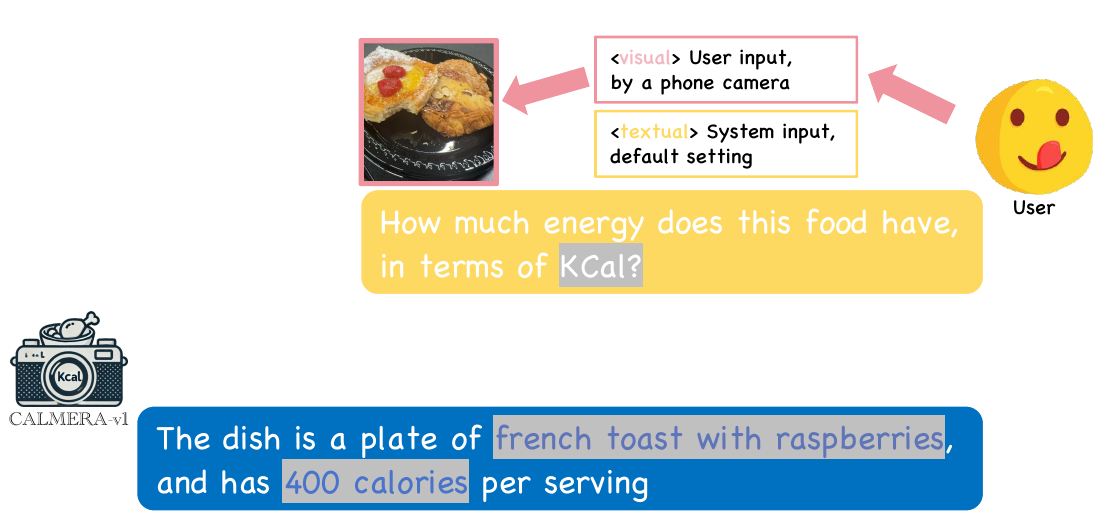}
    \caption{A user interface example of Caloraify}
    \label{fig:intro_teaser}
\end{figure}

Traditional methods for calorie estimation from food images have followed a multi-step pipeline involving food classification, portion size estimation, and caloric calculation \cite{Image-Based-Estimation, CalorieMe, 7410503, liang2017computervisionbasedfoodcalorie}. These approaches often rely on reference objects of known size \cite{Image-Based-Estimation} or depth information \cite{7410503} to estimate portion sizes. While effective under controlled conditions, these methods face several limitations. First, the reliance on specific metadata, such as reference objects or depth images, makes them impractical for general users. Second, the multi-module nature of traditional pipelines introduces significant error propagation, as tasks like segmentation, classification, and volume estimation are handled separately. Lastly, hardware dependencies, such as multi-view camera setups \cite{liang2017computervisionbasedfoodcalorie}, increase system complexity and cost, limiting their scalability for mobile or resource-constrained environments.

Recent advancements in open-sourced multi-modal large language models (LLMs) and vision-language models (VLMs), such as LLAVA \cite{llava} and MiniGPT-4 \cite{minigpt4}, have transformed human interaction with AI tools. By integrating visual and textual understanding, these models enable applications ranging from visual question answering (VQA) to generating detailed, context-aware textual responses. LLAVA excels in interactive VQA, allowing users to ask nuanced questions about images, while MiniGPT-4 enhances multi-modal generative capabilities, providing coherent textual outputs based on image inputs.

In vertical domains, LLAVA-Chef \cite{LLaVA-Chef} demonstrates the potential of fine-tuned VLMs for specialized tasks such as food analysis. It combines visual inputs (e.g., food images) with textual data (e.g., dish titles, ingredient lists) to generate recipes, showcasing the ability of VLMs to integrate diverse modalities for domain-specific applications. However, LLAVA-Chef relies on multi-modal inputs with specified ingredient details, limiting its utility when only visual data is available. Furthermore, its focus on recipe generation lacks the precision required for calorie estimation or ingredient quantification, essential for dietary management.

% Our work addresses these limitations by leveraging advanced VLMs to provide accurate calorie estimation and ingredient analysis based entirely on visual inputs. This approach enhances accessibility and practicality, enabling mobile-friendly solutions for real-world dietary monitoring and nutritional analysis.

% Despite the growing demand for tools to manage calorie intake, existing calorie estimation systems  are often hindered by outdated methodologies, limited accuracy, and the need for specific and complex input data. These limitations render such solutions impractical for real-world applications, particularly for diverse user bases.

% Despite widespread recognition of this issue, effective tools to manage calorie intake—an essential step in combating obesity—remain limited. Many existing solutions rely on outdated data and simplistic methodologies, rendering them ineffective in diverse and realistic dietary scenarios. Furthermore, several of these approaches depend on specialized hardware, such as depth sensors or multi-view imaging systems, which are inaccessible to most users, particularly in resource-limited settings. Traditional deep learning-based methods often incorporate multi-module pipelines that are prone to error propagation and inconsistent optimization, further diminishing their reliability.

% 3. Proposed methods
% data curation + framework design + training
% contribution summary

To address the aforementioned limitations, we present \textbf{CaLoRAify}, a novel VLM framework for accurate calorie estimation and ingredient analysis. By leveraging visual-text pairing during training, our system enables users to perform inference with only a single monocular food image as input (shown in Fig. \ref{fig:intro_teaser}) to obtain detailed ingredient analysis and calorie estimation.
Unlike previous works that require inputs with specific data formats (e.g. food ingredients, depth information) from users, our approach simplifies the process by requiring only a single food image at inference time while retaining the flexibility of agent-based conversational interaction. This enables users to engage with the system through natural queries for more detailed explanations or clarifications, highlighting the advantages of vision-language models in real-world dietary applications. 

We curated a domain-specific dataset, \textbf{CalData}, consisting of 330K \textbf{image-text} pairs derived from Recipe1M+ \cite{recipe1m} and augmented with nutrition facts. This dataset supports fine-tuning VLMs for tasks like ingredient identification and caloric prediction. 
To enhance the system’s performance in this specialized domain, we proposed a novel training pipeline that combines \textbf{Low-rank Adaptation (LoRA)} \cite{LoRA} with \textbf{Retrieval-Augmented Generation (RAG)} \cite{Lewis2020RetrievalAugmentedGF} techniques: we first tune the model to identify ingredients and their quantities from a given image, then with these predictions, we leverage RAG to query an external vector database, retrieving accurate nutritional information for precise calorie estimation.
To augment the training data, we employed a rephrasing model \cite{Yao_2024} to diversify the question set from a base question. Through end-to-end training and the integration of RAG techniques, \textbf{CaLoRAify} improved VLMs' performance in vertical domain applications in personalized dietary management. \noindent Our major contributions are summarized as follows:
\begin{itemize}
    \item We introduce \textbf{CaLoRAify}, a novel vision-language model (VLM) framework tailored for accurate calorie estimation and ingredient analysis. During inference, our system requires only a monocular food image as user input, while internally leveraging visual-text alignment to support calorie estimation and ingredient recognition.

    \item We curate and release a domain-specific dataset, \textbf{CalData}, containing 330K image-text pairs by augmenting Recipe1M+ with detailed nutritional facts. This dataset enables fine-tuning of VLMs for challenging food-specific tasks, such as ingredient identification and calorie prediction.
    
    \item We proposed a novel pipeline for calorie estimation using VLMs, incorporating food ingredient and quantity prediction as an intermediate step, significantly improving system accuracy and computational efficiency for real-world dietary applications.
\end{itemize}

\section{Related Work}
\label{sec:related_work}

We briefly review relevant works on traditional calorie estimation models and multi-modal LLMs related to food.

\textbf{Traditional Calorie Estimation Methods.}
Traditional approaches \cite{Image-Based-Estimation, CalorieMe, 7410503, liang2017computervisionbasedfoodcalorie} for estimating calorie estimation from food images have involved several key steps. First, identify the type of food present in the image. Second, determine the volume or mass of the identified food item. Traditional approaches often utilize reference objects \cite{Image-Based-Estimation} of known size within the image or use depth information \cite{7410503} to estimate portion size. Third, combine the identified food type and estimated portion size to calculate caloric content. These methods, while effective in certain scenarios, exhibit notable limitations. Specific metadata like reference object or depth images are not always present to users. Another challenge comes from the multi-module nature of these pipelines. Tasks such as food segmentation, classification, and volume estimation are often handled in separate stages, leading to error propagation across modules.  Additionally, hardware dependencies, such as the requirement for multi-view camera setups\cite{liang2017computervisionbasedfoodcalorie}, further increase the complexity and cost of implementation, making such systems unsuitable for broader adoption in mobile or resource-constrained settings.

\textbf{Multi-Modal Large Language Models for Food.}
\textbf{Multi-Modal Large Language Models for Food.}
Recent advancements in multimodal large foundation models have demonstrated their capacity to transfer generalizable internet knowledge to specialized downstream tasks \cite{yang2024transferringfoundationmodelsgeneralizable}. Models like LLAVA-Chef \cite{LLaVA-Chef} have introduced new possibilities for food-related applications. LLAVA-Chef leverages multi-modal inputs—including dish titles, ingredient lists, and images—to predict recipes, demonstrating the capability of integrating visual and textual information for food analysis. However, LLAVA-Chef’s focus on predicting recipes based on diverse input modalities makes it less suited for scenarios requiring precise estimation of ingredient amounts or caloric content solely from images. Our approach, by contrast, directly predicts ingredient quantities and calorie estimations based solely on visual inputs, making it more suitable for convenient and real-world applications, such as mobile-based food calorie tracking. 

\textbf{Tuning Strategies for LLMs.}
Fine-tuning strategies have proven essential in adapting large language models for specific tasks. Techniques like low-rank adaptation (LoRA) \cite{LoRA} and prompt tuning focus on optimizing specific layers, enabling efficient adaptation to new tasks without requiring extensive computational resources. Recent studies, such as LIMA\cite{zhou2023limaalignment}, demonstrate that minimal task-specific data can effectively refine pre-trained models while preserving their general reasoning capabilities.  However, unlike traditional text-only applications, the Vision-Language Reasoning (VLR) tasks must align visual and textual information seamlessly, posing unique challenges for fine-tuning strategies. In this paper, we aim to explore whether the findings from LIMA remain valid, particularly when adapting models like MiniGPT-4v2\cite{Chen2023MiniGPTv2LL} for calorie estimation tasks.

\section{CalData Dataset}
\label{sec:data}

For this study, we curated an open-source comprehensive dataset (CalData) tailored to the task of food calorie estimation. The dataset was derived by combining multiple sources, including a large-scale receipt dataset (1M+ entries) and a nutrition instruction dataset containing detailed food amounts. Following a class-balanced sampling strategy \cite{Yao_2023_ICCV}, we identified 5801 unique samples. Each sample was associated with multiple images, resulting in an initial pool of 76,767 images.

For training, we organized the dataset hierarchically for efficient management. Each sample was paired with a maximum of five representative images and combined with five manually constructed instruction sets, creating five image-text pairs per sample. We split the dataset within each sample (recipe) into training (191,433 pairs), validation (63,811 pairs), and test (63,811 pairs) sets. This structured approach ensures robust training and evaluation for vision-language tasks in food calorie estimation.
\section{Method}
\label{sec:method}

\begin{figure*}[h]
    \centering
    \includegraphics[width=0.8\linewidth]{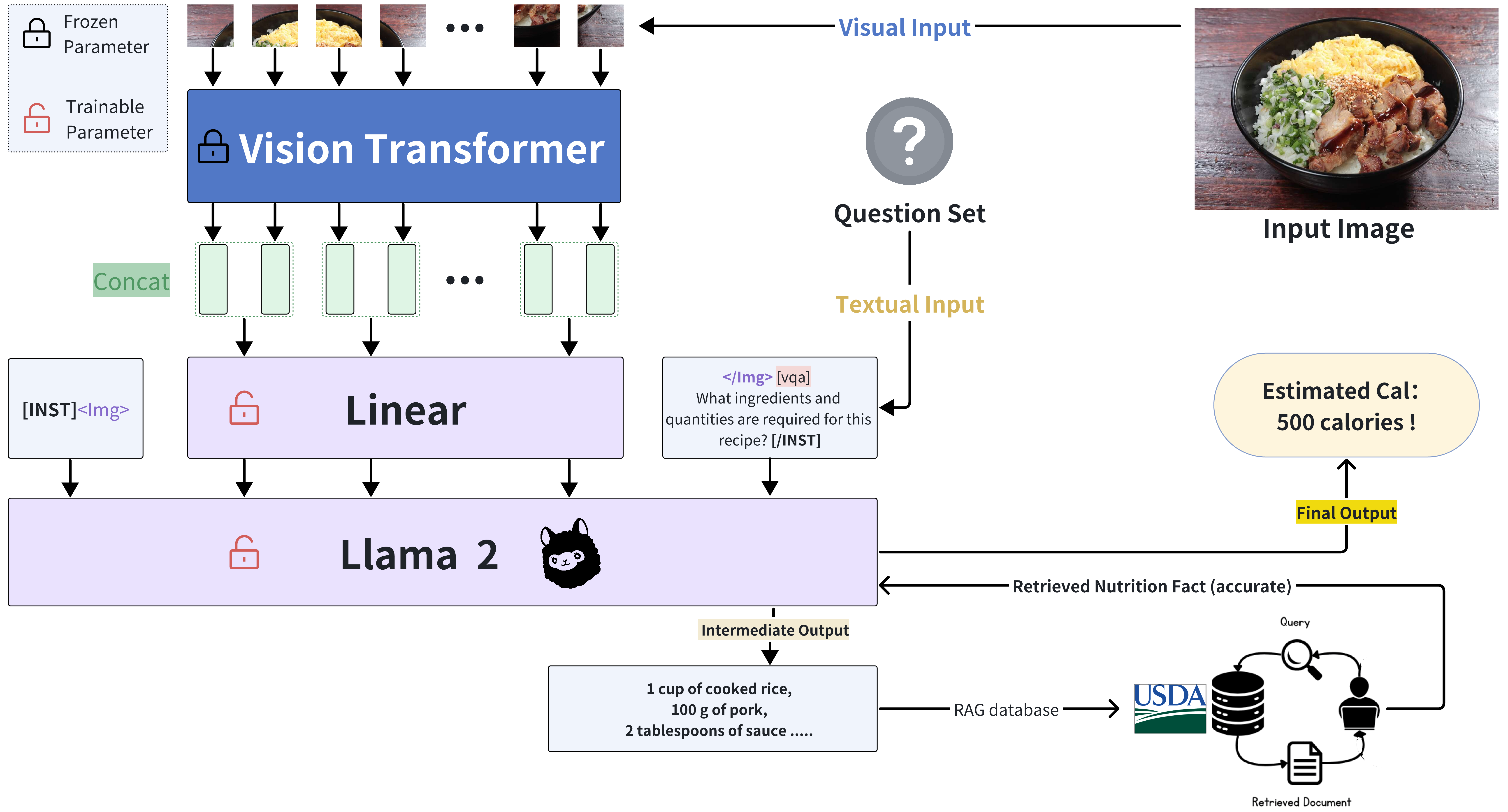}
    \caption{The workflow is similar to \cite{Chen2023MiniGPTv2LL}, beginning with the pre-trained Vision Transformer (ViT) processing the input dish image to extract tokenized visual representations, which capture key features of the dish. Guided by the [vqa] identifier, the LLaMA-2 module formulates a structured question, such as “What ingredients and quantities are required for this recipe?”, to direct subsequent tasks. This query is sent to the Retrieval-Augmented Generation (RAG) module, which retrieves relevant information, including ingredients and their nutritional values, from an external database. Finally, LLaMA-2 integrates the retrieved text and visual features to generate comprehensive outputs, such as ingredient quantities and calorie estimates, presented in an interpretable format.}
    \label{fig:method_framework}
\end{figure*}

With the rapid development of vision language models and large language models, more and more research reveals their strong ability to understand images, enabling them to focus on key aspects just like humans. Most importantly, these large models have well-trained backbones that give them excellent generalization abilities, allowing them to interpret images they have never encountered during training without requiring few-shot training \cite{Hou2019CrossAN, Lai2023ClusteredpatchEC} for each test image. However, there can be instances of hallucination phenomena in the references provided. To address this, we decided to combine the vision language model with the recovered generation (RAG) to build a standard nutrition facts table aiming for more accurate results.
\subsection{Preliminaries}

\subsubsection{MiniGPT4}
MiniGPT4 \cite{Chen2023MiniGPTv2LL}, also known as MiniGPT-v2 is a vision-language model designed as a unified interface for various vision-language tasks, such as image description, visual question answering (VQA), and visual grounding. It achieves this by leveraging a large language model (LLaMA-2 \cite{Touvron2023Llama2O}) as a backbone and aligning it with visual inputs processed by a Vision Transformer (ViT \cite{Dosovitskiy2020AnII}). MiniGPT-v2 uses Task-Specific Identifiers: Unique tokens are introduced for each task (e.g., [vqa], [grounding]) to reduce ambiguity and improve task performance. Here, we choose [vqa] as the task-specific identifier of the model to perform the ingredient detection and calorie estimation. To achieve multimodality, the visual inputs from the frozen ViT backbone are downsampled by concatenating adjacent tokens, improving computational efficiency without compromising resolution. MiniGPT4 followed a three-stage training strategy to train each module individually and was trained on a mix of weakly-labeled datasets, fine-grained datasets, and multi-modal instruction datasets, progressively enhancing its multi-tasking and conversational abilities. It achieves state-of-the-art results and performs comparably to other generalist models on benchmarks like RefCOCO \cite{Chen2024RevisitingRE}, OKVQA \cite{Schwenk2022AOKVQAAB}, and GQA \cite{Hudson2019GQAAN}, excelling in tasks requiring grounded visual understanding.

\subsubsection{Retrieval-Augmented Generation (RAG)}
Retrieval-Augmented Generation (RAG) \cite{Lewis2020RetrievalAugmentedGF, Ma2023QueryRF, Asai2023SelfRAGLT} is a hybrid model architecture designed to enhance the performance of knowledge-intensive tasks by combining pre-trained parametric memory with nonparametric memory. Parametric memory consists of a pre-trained seq2seq transformer model, such as BART \cite{Lewis2019BARTDS}, which handles the language vectorization. In contrast, the non-parametric memory is implemented using a dense vector index of external knowledge sources like Wikipedia, accessed via a pre-trained neural retriever (e.g., Dense Passage Retriever \cite{Karpukhin2020DensePR}).

% RAG models operate by retrieving relevant documents or text passages based on the input query and using these retrieved pieces as additional context during language generation. The retrieval process treats the documents as latent variables, with two main variants of RAG: RAG-Sequence, which generates the entire output sequence conditioned on a single document, and RAG-Token, which allows the model to retrieve different documents for each token in the output. Both components—the retriever and generator—are trained jointly in an end-to-end manner, enabling the model to learn to access and leverage external knowledge effectively.

RAG has demonstrated state-of-the-art performance on tasks like open-domain question answering, abstractive summarization, and fact verification, outperforming purely parametric models by generating more specific, factual, and diverse responses. Additionally, its non-parametric memory can be dynamically updated, offering flexibility and interoperability. In our work, to obtain the precise calorie estimation, we seek reference from a database assembled by the United States Department of Agriculture (USDA) \cite{usda}.

\subsection{CaLoRAify}

Our proposed CaLoRAify framework integrates MiniGPT-v2 with a Retrieval-Augmented Generation (RAG) mechanism to address vision-language tasks involving ingredient recognition and calorie estimation. Inspired by the architecture of MiniGPT-v2, we leverage its modular design combining a Vision Transformer (ViT) and a large language model (LLaMA-2), adapted for multimodal tasks. This integration ensures robust performance in calorie estimation, enhanced by external database retrieval for accurate results.

As shown in Fig \ref{fig:method_framework}, input image patches are fed to a frozen ViT encoder to extract visual features. We then linearly project these features into the LLaMA-2 embedding space after being concatenated to reduce dimensionality, which improves computational efficiency while preserving key image details. The task-specific identifier [vqa] directs the system to interpret the query for ingredient recognition.

 After receiving the ingredient detection from the LLM, a query related to the nutritional information of the dish is processed using the RAG module. Here we use Sentence Transformers \cite{Reimers2019SentenceBERTSE} as the text encoder to generate queries in the text embedding space. The query then retrieves relevant documents from the USDA nutritional data database. The retrieved textual content, such as the quantities of ingredients and their caloric values, is concatenated with the LLaMA-2 embeddings with the instruction. This combination enables the model to generate an accurate calorie estimation while mitigating hallucinations.
\section{Experiment}
\label{sec:exp}

\subsection{Experimental Setup}

The experiments were conducted on a server equipped with four NVIDIA A800 GPUs, each with 80GB of memory. The training process took approximately 7 hours under these conditions. To maximize GPU memory utilization, the batch size was set to 12 per GPU. The training consists of 8 epochs, with each epoch having 875 iterations. The initial learning rate was configured as $10^{-5}$ and we used \verb|LinearWarmupCosineLRScheduler| as the learning rate scheduler. The LoRA parameters were configured with a rank of 64 and an alpha value of 16.

\subsection{Input Format}

The input format of our model is as follows:
\begin{align*}
\textit{[INST]$<$Img$><$ImageFeature$><$/Img$>$}\\
\textit{[Task Identifier] Instruction [/INST]}
\end{align*}
where we structure the user input into two parts. The first part is image features and the second part is the instruction input. Here we assign the Task Identifier as 
Visual Question Answering (VQA).

\subsection{Metrics Results}

\begin{table}[h!]
    \centering
    \begin{tabular}{lccc}
        \toprule
        \textbf{Metric} & \textbf{Baseline} & \textbf{Fine-tune} & \textbf{Increase \%} \\
        \midrule
        ROUGE-1 & 0.209  & 0.2173 & 3.97\% \\
        ROUGE-2 & 0.0611 & 0.0947 & 55.01\% \\
        ROUGE-L & 0.1643 & 0.1734 & 5.53\% \\
        ROUGE-Lsum & 0.1643 & 0.1733 & 5.48\% \\
        BLEU & 0.0135 & 0.0218 & 61.48\% \\
        SacreBLEU & 1.3518 & 2.1845 & 61.60\% \\
        BERTScore (P) & 0.8441 & 0.846 & 0.23\% \\
        BERTScore (R) & 0.8117 & 0.8135 & 0.22\% \\
        BERTScore (F1) & 0.8273 & 0.8289 & 0.19\% \\
        Aggregate Metrics & 0.431 & 0.4662 & 8.16\% \\
        \bottomrule
    \end{tabular}
    \caption{Performance comparison between Baseline and Fine-tune models.}
    \label{tab:result}
\end{table}

Table \ref{tab:result} shows the metrics results of our experiment, where the baseline model is before fine-tuning, with the backbone of MiniGPT-4 \cite{minigpt4}, and the fine-tuned is the model we trained based on the baseline. 

The Aggregate Metrics is defined as the weighted average of ROUGE-L and BLEU: 
$$
L_{agg} = \lambda_{rouge}L_{rouge} + \lambda_{bleu}L_{bleu}
$$
From the metrics in Table \ref{tab:result} we show that our fine-tuning improves the accuracy of the model. 

\subsection{Qualitative Results}

\begin{figure}[h]
    \centering
    \includegraphics[width=1.05\linewidth]{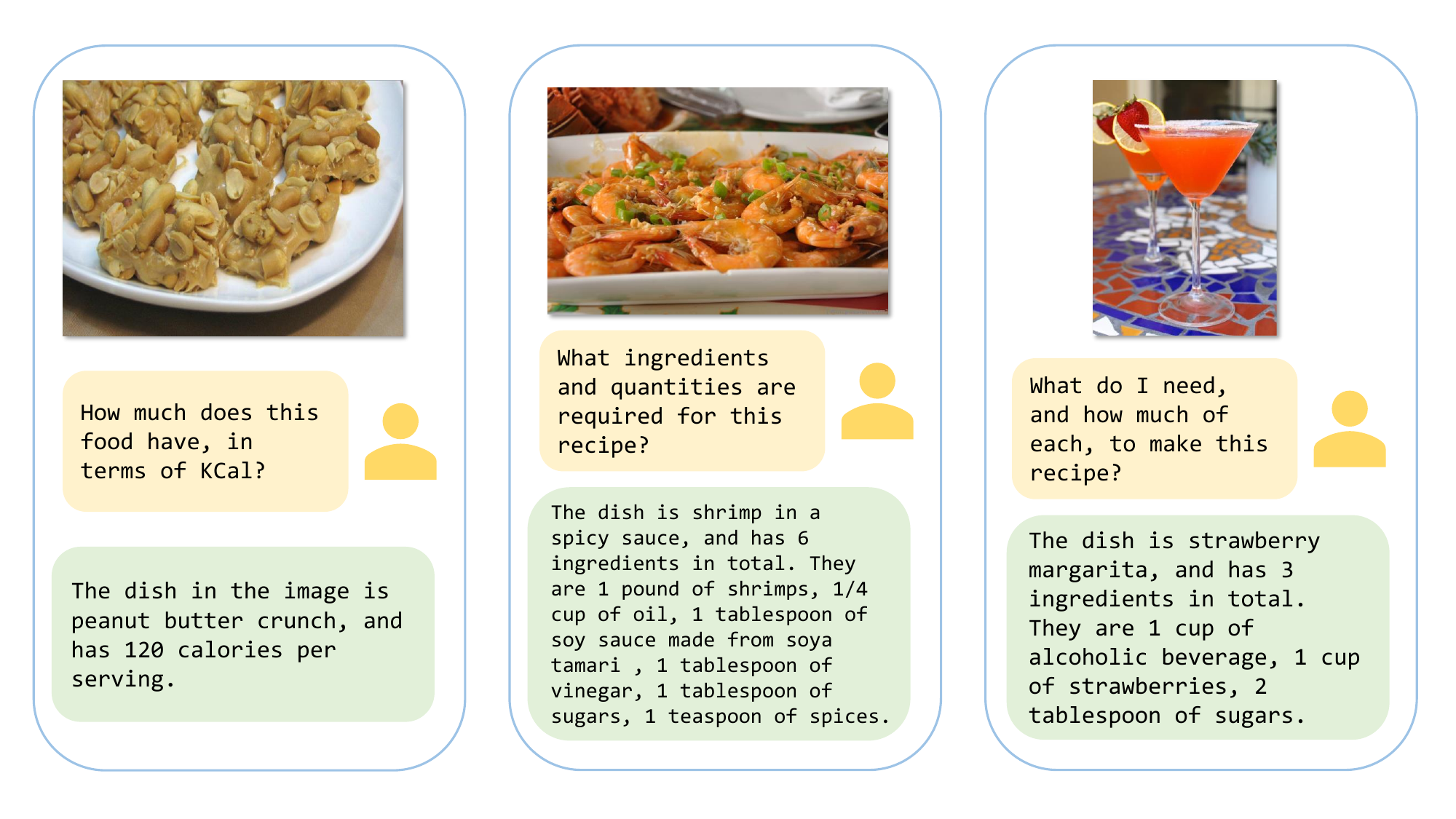}
    \caption{Qualitative results of the model output}
    \label{fig:demo}
\end{figure}

In Figure \ref{fig:demo} we show some examples of our model performing different food-related VQA tasks.

\section{Conclusion and Future Work}
\label{sec:conclusion}

In this paper, we introduced CaLoRAify, a novel framework for accurate calorie estimation and ingredient analysis using vision-language models. By leveraging the capabilities of MiniGPT-v2 and enhancing them with Retrieval-Augmented Generation (RAG) and Low-Rank Adaptation (LoRA), our approach demonstrated significant improvements in both accuracy and computational efficiency for food-specific tasks. Through the creation of our domain-specific dataset CalData, consisting of 330K image-text pairs, we enabled fine-tuning of vision-language models for real-world vertical domain applications, achieving robust performance in calorie estimation tasks without requiring complex multi-step pipelines.

Our experimental results highlight the effectiveness of our method in addressing key limitations of traditional calorie estimation approaches, such as error propagation and hardware dependencies. CaLoRAify streamlines the process by requiring only a single food image during inference, while still supporting flexible, agent-based conversational interactions. The integration of RAG further ensures that calorie estimations are grounded in accurate and up-to-date external knowledge, mitigating hallucinations often observed in purely generative models.

For future work, we aim to explore several directions to further enhance the system:

\begin{itemize}
    \item \textbf{Real-Time Inference}: Optimizing the framework for deployment on mobile and edge devices to enable real-time calorie tracking in everyday scenarios without uploading the image to the cloud.
    \item \textbf{Expanded Datasets}: Incorporating additional datasets, such as regional and cultural food databases, to improve the model’s generalization to diverse cuisines.
    \item \textbf{Interactive Features}: Developing more interactive features, such as recipe generation or personalized dietary recommendations, based on user-specific goals or constraints.
\end{itemize}

By addressing these future directions, we aim to further refine CaLoRAify as a versatile and practical tool for dietary management and beyond, pushing the boundaries of vision-language applications in specialized domains.

% Extra: we plan to use a sentence transformer as 
% text encoder
{
    \small
    \bibliographystyle{ieeenat_fullname}
    \bibliography{main}

\begin{thebibliography}{32}
\providecommand{\natexlab}[1]{#1}
\providecommand{\url}[1]{\texttt{#1}}
\expandafter\ifx\csname urlstyle\endcsname\relax
  \providecommand{\doi}[1]{doi: #1}\else
  \providecommand{\doi}{doi: \begingroup \urlstyle{rm}\Url}\fi

\bibitem[Cal()]{CalAI_Revenue}
Ai-based calorie estimation app calai reports \$50 million revenue.
\newblock \url{https://www.businessinsider.com/calai-revenue-growth-2023}.
\newblock Accessed: 2024-10-11.

\bibitem[Los()]{LoseIt_Success}
Loseit app reaches new milestone in calorie tracking.
\newblock \url{https://www.loseit.com/news/milestone}.
\newblock Accessed: 2024-10-11.

\bibitem[MyF()]{MyFitnessPal_UserBase}
Myfitnesspal user statistics.
\newblock \url{https://www.myfitnesspal.com/statistics}.
\newblock Accessed: 2024-10-11.

\bibitem[NIH()]{NIH}
National institutes of health overweight \& obesity statistics.
\newblock \url{https://www.niddk.nih.gov/health-information/health-statistics/overweight-obesity}.
\newblock Accessed: 2024-10-11.

\bibitem[usd()]{usda}
Usda food calorie database.
\newblock \url{https://fdc.nal.usda.gov/fdc-app.html}.
\newblock Accessed: 2024-10-10.

\bibitem[Asai et~al.(2023)Asai, Wu, Wang, Sil, and Hajishirzi]{Asai2023SelfRAGLT}
Akari Asai, Zeqiu Wu, Yizhong Wang, Avirup Sil, and Hannaneh Hajishirzi.
\newblock Self-rag: Learning to retrieve, generate, and critique through self-reflection.
\newblock \emph{ArXiv}, abs/2310.11511, 2023.

\bibitem[Chen et~al.(2023)Chen, Zhu, et~al.]{Chen2023MiniGPTv2LL}
Jun Chen, Deyao Zhu, et~al.
\newblock Minigpt-v2: large language model as a unified interface for vision-language multi-task learning.
\newblock \emph{ArXiv}, abs/2310.09478, 2023.

\bibitem[Chen et~al.(2024)Chen, Wei, Zhao, Song, Wu, Peng, Chan, and Zhang]{Chen2024RevisitingRE}
Jierun Chen, Fangyun Wei, Jinjing Zhao, Sizhe Song, Bohuai Wu, Zhuoxuan Peng, S.-H.~Gary Chan, and Hongyang Zhang.
\newblock Revisiting referring expression comprehension evaluation in the era of large multimodal models.
\newblock \emph{ArXiv}, abs/2406.16866, 2024.

\bibitem[Dosovitskiy et~al.(2020)Dosovitskiy, Beyer, Kolesnikov, Weissenborn, Zhai, Unterthiner, Dehghani, Minderer, Heigold, Gelly, Uszkoreit, and Houlsby]{Dosovitskiy2020AnII}
Alexey Dosovitskiy, Lucas Beyer, Alexander Kolesnikov, Dirk Weissenborn, Xiaohua Zhai, Thomas Unterthiner, Mostafa Dehghani, Matthias Minderer, Georg Heigold, Sylvain Gelly, Jakob Uszkoreit, and Neil Houlsby.
\newblock An image is worth 16x16 words: Transformers for image recognition at scale.
\newblock \emph{ArXiv}, abs/2010.11929, 2020.

\bibitem[Ege et~al.(2019)Ege, Ando, Tanno, Shimoda, and Yanai]{Image-Based-Estimation}
Takumi Ege, Yoshikazu Ando, Ryosuke Tanno, Wataru Shimoda, and Keiji Yanai.
\newblock Image-based estimation of real food size for accurate food calorie estimation.
\newblock In \emph{2019 IEEE Conference on Multimedia Information Processing and Retrieval (MIPR)}, pages 274--279, 2019.

\bibitem[Hou et~al.(2019)Hou, Chang, Ma, Shan, and Chen]{Hou2019CrossAN}
Rui Hou, Hong Chang, Bingpeng Ma, S. Shan, and Xilin Chen.
\newblock Cross attention network for few-shot classification.
\newblock In \emph{Neural Information Processing Systems}, 2019.

\bibitem[Hu et~al.(2021)Hu, Shen, Wallis, Allen-Zhu, Li, Wang, and Wang]{LoRA}
Edward~J. Hu, Yelong Shen, Phillip Wallis, Zeyuan Allen-Zhu, Yuanzhi Li, Shean Wang, and Lu Wang.
\newblock Lora: Low-rank adaptation of large language models.
\newblock \emph{arXiv preprint arXiv:2106.09685}, 2021.

\bibitem[Hudson and Manning(2019)]{Hudson2019GQAAN}
Drew~A. Hudson and Christopher~D. Manning.
\newblock Gqa: A new dataset for real-world visual reasoning and compositional question answering.
\newblock \emph{2019 IEEE/CVF Conference on Computer Vision and Pattern Recognition (CVPR)}, pages 6693--6702, 2019.

\bibitem[Karpukhin et~al.(2020)Karpukhin, Oğuz, Min, Lewis, Wu, Edunov, Chen, and tau Yih]{Karpukhin2020DensePR}
Vladimir Karpukhin, Barlas Oğuz, Sewon Min, Patrick Lewis, Ledell~Yu Wu, Sergey Edunov, Danqi Chen, and Wen tau Yih.
\newblock Dense passage retrieval for open-domain question answering.
\newblock \emph{ArXiv}, abs/2004.04906, 2020.

\bibitem[Lai et~al.(2023)Lai, Yang, Zhou, et~al.]{Lai2023ClusteredpatchEC}
Jinxiang Lai, Siqian Yang, Junhong Zhou, et~al.
\newblock Clustered-patch element connection for few-shot learning.
\newblock In \emph{International Joint Conference on Artificial Intelligence}, 2023.

\bibitem[Lewis et~al.(2019)Lewis, Liu, et~al.]{Lewis2019BARTDS}
Mike Lewis, Yinhan Liu, et~al.
\newblock Bart: Denoising sequence-to-sequence pre-training for natural language generation, translation, and comprehension.
\newblock In \emph{Annual Meeting of the Association for Computational Linguistics}, 2019.

\bibitem[Lewis et~al.(2020)Lewis, Perez, Piktus, et~al.]{Lewis2020RetrievalAugmentedGF}
Patrick Lewis, Ethan Perez, Aleksandara Piktus, et~al.
\newblock Retrieval-augmented generation for knowledge-intensive nlp tasks.
\newblock \emph{ArXiv}, abs/2005.11401, 2020.

\bibitem[Liang and Li(2017)]{liang2017computervisionbasedfoodcalorie}
Yanchao Liang and Jianhua Li.
\newblock Computer vision-based food calorie estimation: dataset, method, and experiment, 2017.

\bibitem[Liu et~al.(2023)Liu, Li, Wu, and Lee]{llava}
Haotian Liu, Chunyuan Li, Qingyang Wu, and Yong~Jae Lee.
\newblock Visual instruction tuning, 2023.

\bibitem[Ma et~al.(2023)Ma, Gong, He, Zhao, and Duan]{Ma2023QueryRF}
Xinbei Ma, Yeyun Gong, Pengcheng He, Hai Zhao, and Nan Duan.
\newblock Query rewriting for retrieval-augmented large language models.
\newblock \emph{ArXiv}, abs/2305.14283, 2023.

\bibitem[Magid et~al.(2023)Magid, Ibrahim, Kawashti, Mohamed, Sabry, Hindy, Khaled, and Mohamed]{CalorieMe}
Bavlly Magid, Mohamed Ibrahim, Yomna~A. Kawashti, Mazen Mohamed, Mohamed Sabry, Hanan Hindy, Mazen Khaled, and Waleed Mohamed.
\newblock Calorieme: An image-based calorie estimator system.
\newblock In \emph{2023 Eleventh International Conference on Intelligent Computing and Information Systems (ICICIS)}, pages 555--560, 2023.

\bibitem[Marin et~al.(2018)Marin, Karp, Parikh, and Farhadi]{recipe1m}
J. Marin, P. Karp, D. Parikh, and A. Farhadi.
\newblock Recipe1m+: A dataset for learning cross-modal embeddings for cooking recipes and food images.
\newblock \emph{arXiv preprint arXiv:1810.06553}, 2018.

\bibitem[Mohbat and Zaki(2024)]{LLaVA-Chef}
Fnu Mohbat and Mohammed~J. Zaki.
\newblock Llava-chef: A multi-modal generative model for food recipes.
\newblock In \emph{Proceedings of the 33rd ACM International Conference on Information and Knowledge Management}, page 1711–1721. ACM, 2024.

\bibitem[Myers et~al.(2015)Myers, Johnston, Rathod, Korattikara, Gorban, Silberman, Guadarrama, Papandreou, Huang, and Murphy]{7410503}
Austin Myers, Nick Johnston, Vivek Rathod, Anoop Korattikara, Alex Gorban, Nathan Silberman, Sergio Guadarrama, George Papandreou, Jonathan Huang, and Kevin Murphy.
\newblock Im2calories: Towards an automated mobile vision food diary.
\newblock In \emph{2015 IEEE International Conference on Computer Vision (ICCV)}, pages 1233--1241, 2015.

\bibitem[Reimers et~al.(2019)]{Reimers2019SentenceBERTSE}
Nils Reimers et~al.
\newblock Sentence-bert: Sentence embeddings using siamese bert-networks.
\newblock In \emph{Conference on Empirical Methods in Natural Language Processing}, 2019.

\bibitem[Schwenk et~al.(2022)Schwenk, Khandelwal, Clark, Marino, and Mottaghi]{Schwenk2022AOKVQAAB}
Dustin Schwenk, Apoorv Khandelwal, Christopher Clark, Kenneth Marino, and Roozbeh Mottaghi.
\newblock A-okvqa: A benchmark for visual question answering using world knowledge.
\newblock In \emph{European Conference on Computer Vision}, 2022.

\bibitem[Touvron and others(2023)Touvron et~al.]{Touvron2023Llama2O}
Hugo Touvron and Louis~Martin others.
\newblock Llama 2: Open foundation and fine-tuned chat models.
\newblock \emph{ArXiv}, abs/2307.09288, 2023.

\bibitem[Yang et~al.(2024)Yang, Tan, Jin, Yao, Liu, Fu, Song, Wu, and Wang]{yang2024transferringfoundationmodelsgeneralizable}
Jiange Yang, Wenhui Tan, Chuhao Jin, Keling Yao, Bei Liu, Jianlong Fu, Ruihua Song, Gangshan Wu, and Limin Wang.
\newblock Transferring foundation models for generalizable robotic manipulation, 2024.

\bibitem[Yao and Li(2023)]{Yao_2023_ICCV}
Dongyu Yao and Boheng Li.
\newblock Dual-level interaction for domain adaptive semantic segmentation.
\newblock In \emph{Proceedings of the IEEE/CVF International Conference on Computer Vision (ICCV) Workshops}, pages 4527--4536, 2023.

\bibitem[Yao et~al.(2024)Yao, Zhang, Harris, and Carlsson]{Yao_2024}
Dongyu Yao, Jianshu Zhang, Ian~G. Harris, and Marcel Carlsson.
\newblock Fuzzllm: A novel and universal fuzzing framework for proactively discovering jailbreak vulnerabilities in large language models.
\newblock In \emph{ICASSP 2024 - 2024 IEEE International Conference on Acoustics, Speech and Signal Processing (ICASSP)}, page 4485–4489. IEEE, 2024.

\bibitem[Zhou et~al.(2023)Zhou, Liu, Xu, Iyer, Sun, Mao, Ma, Efrat, Yu, Yu, Zhang, Ghosh, Lewis, Zettlemoyer, and Levy]{zhou2023limaalignment}
Chunting Zhou, Pengfei Liu, Puxin Xu, Srini Iyer, Jiao Sun, Yuning Mao, Xuezhe Ma, Avia Efrat, Ping Yu, Lili Yu, Susan Zhang, Gargi Ghosh, Mike Lewis, Luke Zettlemoyer, and Omer Levy.
\newblock Lima: Less is more for alignment, 2023.

\bibitem[Zhu et~al.(2023)Zhu, Chen, Li, Zhang, Zhao, Wei, Wu, Zhang, Fu, Cao, et~al.]{minigpt4}
Deyao Zhu, Jun Chen, Xiaoqian Li, Guangsheng Zhang, Jianfei Zhao, Yixiao Wei, Yuhao Wu, Lijun Zhang, Yanwei Fu, Yuanliang Cao, et~al.
\newblock Minigpt-4: Enhancing vision-language understanding with advanced large language models.
\newblock \emph{arXiv preprint arXiv:2304.10592}, 2023.

\end{thebibliography}
}


\end{document}